\documentclass[conference]{IEEEtran}
\IEEEoverridecommandlockouts
\usepackage{cite}
\usepackage{amsmath,amssymb,amsfonts}
\usepackage{algorithmic}
\usepackage{textcomp}
\usepackage{times}
\usepackage[none]{hyphenat}
\usepackage{url}
\usepackage{latexsym}
\usepackage{array}
\usepackage{lipsum}
\usepackage{pgfplots}
\usepackage{booktabs}
\usepackage{multirow}
\usepackage{xcolor}
\usepackage{hyperref}
\hypersetup{colorlinks = true,
            linkcolor = .,
            urlcolor  = purple,
            citecolor = green,
            anchorcolor = blue}

\usepackage{subfig}
\usepackage{graphicx}
\graphicspath{{images/}}
\newcommand\blfootnote[1]{%
  \begingroup
  \renewcommand\thefootnote{}\footnote{#1}%
  \addtocounter{footnote}{-1}%
  \endgroup
}

\begin{document}

\title{Towards Fine-grained Image Classification with Generative Adversarial Networks and Facial Landmark Detection*}
\author{\IEEEauthorblockN{Mahdi Darvish, Mahsa Pouramini, Hamid Bahador \\[0.2cm]}
\IEEEauthorblockA{Department of Electrical  and Computer Science,\\
University of Mohaghegh Ardabili, Ardabil, Iran}}
\maketitle

\begin{abstract}
Fine-grained classification remains a challenging task because distinguishing categories needs learning complex and local differences. Diversity in the pose, scale, and position of objects in an image makes the problem even more difficult. Although the recent Vision Transformer models achieve high performance, they need an extensive volume of input data. To encounter this problem, we made the best use of GAN-based data augmentation to generate extra dataset instances. Oxford-IIIT Pets was our dataset of choice for this experiment. It consists of 37 breeds of cats and dogs with variations in scale, poses, and lighting, which intensifies the difficulty of the classification task. Furthermore, we enhanced the performance of the recent Generative Adversarial Network (GAN), StyleGAN2-ADA model to generate more realistic images while preventing overfitting to the training set. We did this by training a customized version of MobileNetV2 to predict animal facial landmarks; then, we cropped images accordingly. Lastly, we combined the synthetic images with the original dataset and compared our proposed method with standard GANs augmentation and no augmentation with different subsets of training data. We validated our work by evaluating the accuracy of fine-grained image classification on the recent Vision Transformer (ViT) Model. Code is available at: \urlstyle{sf}\href{https://github.com/mahdi-darvish/GANs-augmented-pet-classifier}{\normalfont https://github.com/mahdi-darvish/GAN-augmented-pet-classifier\blfootnote{*This work has been submitted to the IEEE for possible publication.}} \\[0.2cm]
\end{abstract}


\begin{figure}[tb]
    \centering
	\includegraphics[clip,width=0.9\linewidth]{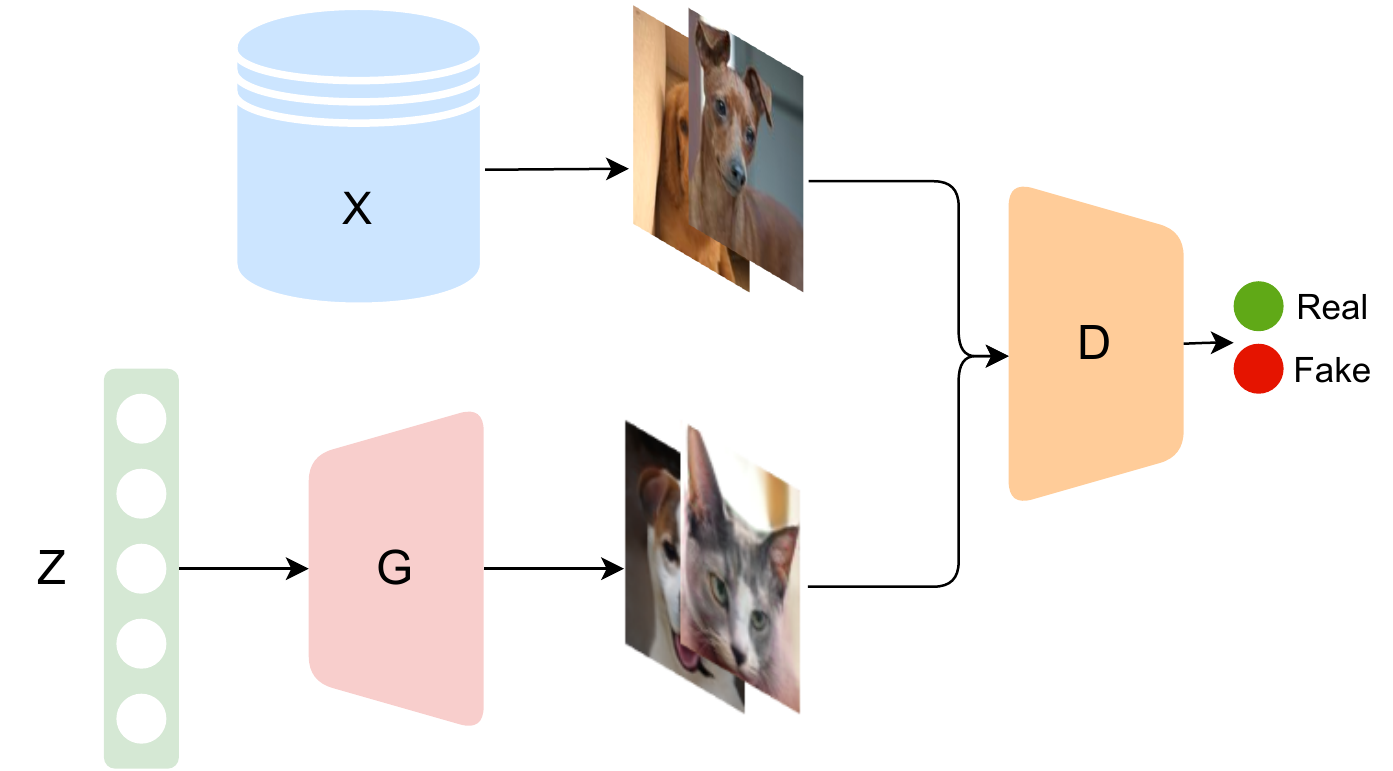}
	\caption{The architecture of Generative Adversarial Network (GAN). The generator (G) takes a random noise (Z) as input to build fake images that would look like real ones (X). The discriminator (D) learns to tell the difference between the generator's fake image from actual data and outputs a probability for the image being real.}
	\label{fig:gans}

\end{figure}

\section{Introduction}
\label{sec:intro}
In recent years, Convolutional Neural Networks and Vision Transformers have grown into the state-of-the-art techniques in computer vision \cite{Sutton2000} with remarkable results on famous datasets, namely ImageNet \cite{5206848}, JFT-300M \cite{Sun_2017_ICCV}. One of the main factors affecting the performance of mentioned algorithms is a considerable quantity of high-resolution and distinct training instances. On the other hand, collecting such datasets requires expertise in the specialized field also is tedious and costly. Moreover, sometimes it is not feasible to collect a satisfactory number of data; for example, in the case of Fine-Grained Classification \cite{Akata_2015_CVPR, DBLP:journals/corr/abs-2106-10587, conde2021exploring}, which is a sub-field of image recognition \cite{He_2016_CVPR}, that tries to differentiate subcategories of a general class. Examples include distinguishing breeds of animals or species of flowers. As a result, several methods have been introduced to tackle these problems. \\

\begin{figure*}[ht]
    \centering
	\includegraphics[clip,width=0.9\linewidth]{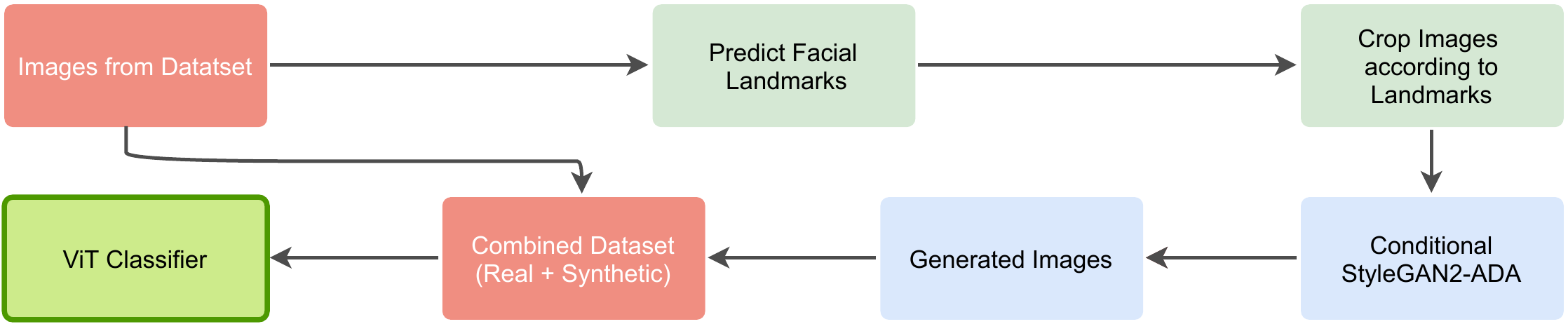}
	\caption{The complete classification pipeline of our method for GAN-based data augmentation. }
	\label{fig:pipe} 
\end{figure*}

Data augmentation techniques have been performed remarkably well as a solution to the problem of limited data \cite{DBLP:journals/corr/abs-1712-04621, jain2021using}. These algorithms use various approaches to incorporate new instances from existing training data which are divided into two groups, traditional methods \cite{zhao2020differentiable, 10.1007/978-3-030-58583-9_34} and image generation using Generative Adversarial Networks (GANs) \cite{goodfellow2014generative}. Color jittering, flipping,  blurring, rotation, and adding a small amount of noise exemplify the former. GANs can produce unseen images with the same statistics as original training data. GANs are capable of training a generative model (automatically finding the patterns from the input data and learning harmonies to produce new examples without supervision) by associating two neural networks, namely, generator and discriminator (see Figure \textcolor{red}{\ref{fig:gans}}). The generator model makes a synthetic example from a random vector. The generated images, then, are fed to the discriminator along with authentic images, and the discriminator attempt to identify the fake ones. These two models are trained together until the success rate of the discriminator stabilized at approximately 50\%. \\

As previously discussed, GANs seem to be an effective method for data augmentation and had exceptional success on certain datasets, e.g.,  100-shot Obama and Panda \cite{zhao2020differentiable}. Still, there are difficulties generalizing to other datasets. The mentioned datasets all have similar characteristics, such as objects (faces) are centered and often have limited distribution. However, it is not always the case; most datasets have objects in various poses and sizes, which leads to poor convergence of GAN models \cite{barnett2018convergence}. As a result, the produced images might have issues with abnormal colors and distribution shifts.\cite{9035107} In other cases, when the size of the training set is not satisfactory, the loss of the discriminator falls rapidly. Still, the validation loss decreases, showing the network just producing images almost identical to the training set. Consequently, when the suggested images add up to the original dataset for computer vision tasks, in particular image classification, the evaluation metrics barely fluctuate, which indicates the impracticality of these methods \cite{sundaram2021ganbased}. \\

To overcome the stated drawbacks, we took advantage of several methods. For this purpose, we experimented with our work on the Oxford-IIIT Pet dataset \cite{6248092}, which consists of different breeds of dogs and cats images. Experimental results demonstrate a noticeable increase in accuracy of highly developed model, Vision Transformer (ViT), \cite{dosovitskiy2021image} on the task of fine-grained image classification. First, we tried generating new realistic samples to feed the model, but produced images were impractical and had poor quality. To solve this problem, we capitalize on a new version of StyleGAN \cite{karras2019stylebased}, the highly developed StyleGAN2-ADA \cite{karras2020training}, which proposed a flexible discriminator augmentation mechanism that improves performance on limited data regimes. As previously suggested, GANs perform adequately when training objects are not centered, so we trained a customized version of the famous CNN architecture, MobileNetV2 \cite{Sandler_2018_CVPR}, to predict facial landmarks \cite{Wu_2018} of cats \cite{weiweizhangandjiansunandxiaooutang2008cat} and dogs \cite{7026060} and cropped instances of the Oxford-IIIT Pet dataset according to predicted landmarks. At last, we trained the StyleGan2-ADA using the cropped images, next off by combining the produced images with the original data, created a more prosperous and extensive dataset. Finally, we trained the upgraded dataset on the ViT benchmark with the same hyper-parameters and achieved a better result. 

This paper commences with introducing some related works in the following section. Next, it explains the approach proposed and presented the experimental setup in sections \textcolor{red}{\ref{sec:method}} and \textcolor{red}{\ref{sec:exper}}, respectively. In section \textcolor{red}{\ref{sec:result}}, the outcomes are reported, and finally, we have a conclusion in section \textcolor{red}{\ref{sec:con}}.

\section{Related Work}

Dataset size is an essential factor in the performance of image classifiers, and the primary issue is overfitting \cite{doi:10.1021/ci0342472, zhao2020differentiable}. When the model is overfitting the training set, it will not generalize well on the test data. Many methods have been proposed over the years to overcome this problem; the most straightforward methods include using regularization \cite{6796297} or dropout \cite{JMLR:v15:srivastava14a}. The former adds a penalty to the norm of the weights so that the results will be less fluctuated. Dropout drops certain connections in the network by removing the percentage of neurons randomly. Later techniques such as batch normalization \cite{cooijmans2017recurrent} re-scale and center each layer, thus making the model more stable and accurate. Transfer learning \cite{10.1007/978-3-030-01424-7_27} is another popular strategy that uses the weights of the same model trained on a different dataset instead of randomly initialize them. All of the suggested approaches are orthogonal to our work, and it is possible to use them all together.

That being the case, it requires specialization and time to tune the parameters to get the best accuracy. Our method generates new images, which data augmentation can apply to and increase the size of the dataset further. 

The most advanced augmentation method is using GANs to generate new never-seen-before samples from current data. Since the invention of GANs\cite{goodfellow2014generative}, various models dependent on GAN were introduced \cite{8667290}. One issue with these models was the lack of ability to distinguish between different classes of training data and no control over generated images. Consequently, Conditional Generative Adversarial Networks (cGAN) \cite{mirza2014conditional} were proposed in 2014, which allowed image generation to be conditional on a possible class label, enabling generate images on a specified target class. Sanfort et al. proved the potential of GAN-based data augmentation \cite{Sandfort2019 } confirmed that using CycleGAN for augmentation in diverse tasks, e.g., segmentation and classification, can substantially boost performance in these domains.

In this paper, we used the state-of-the-art model StyleGAN2-ADA, introduced by Karras et al. in 2020\cite {karras2020training}, which by augmenting the input data perform better than previous ones and better on low data regimes. However, a downside with GAN models is that the objects need to be centered and have similar poses; we study challenging scenarios where mentioned conditions do not apply and overcome this issue by cropping the images from predicted landmarks of samples, in this case, animals, to boost the StyleGAN2-ADA. Furthermore, we showed that these changes not only reduce the Frechet Inception Distance \cite{8110709} score of StyleGAN2-ADA but also, generated images can be used for data augmentation in fine-grained images classification.

\section{Method}
\label{sec:method}
\begin{figure*}[ht]
    \subfloat[\label{subfig-1:dummy}]{%
      \includegraphics[width=0.13\textwidth]{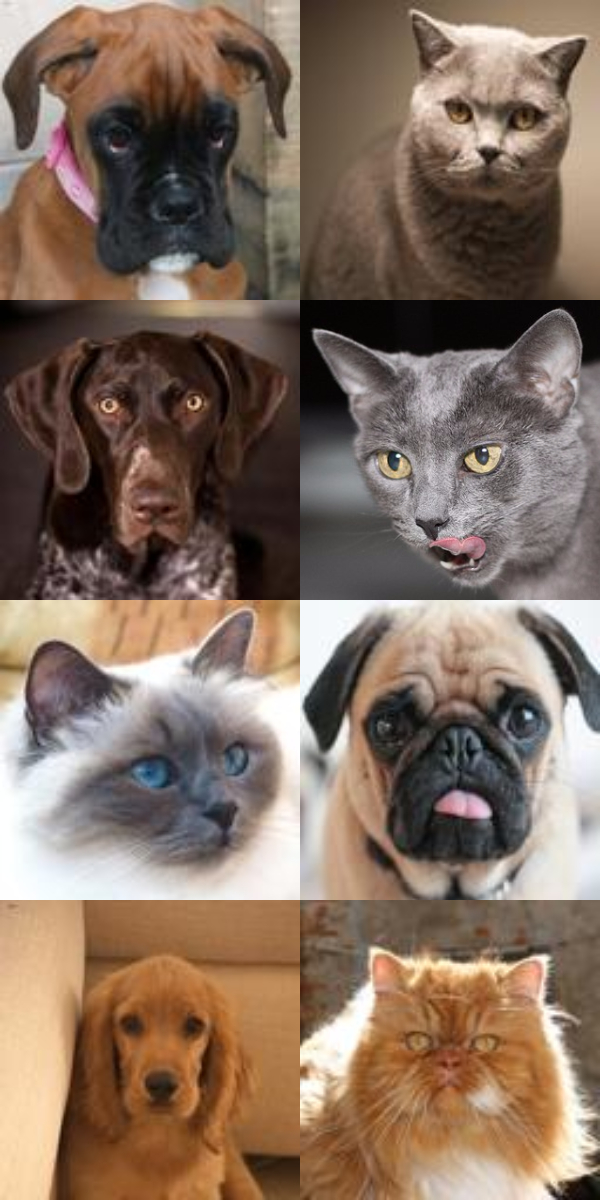}
    }
    \hfill
    \subfloat[\label{subfig-1:dummy}]{%
      \includegraphics[width=0.13\textwidth]{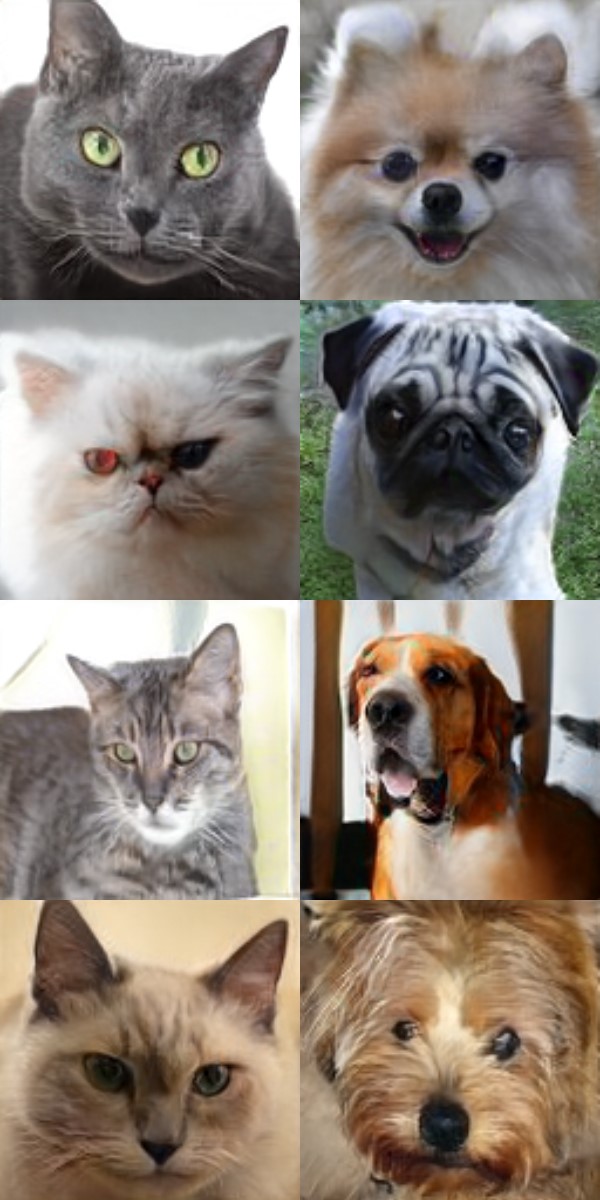}
    }
    \hfill\subfloat[\label{subfig-1:dummy}]{%
      \includegraphics[width=0.13\textwidth]{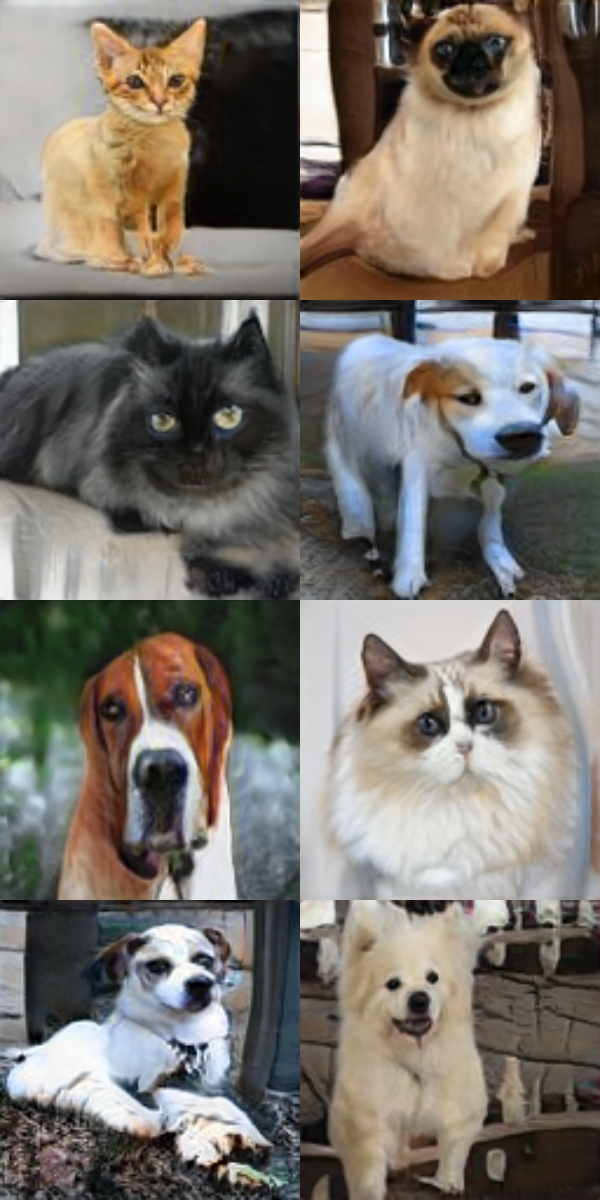}
    }
    \hfill\subfloat[\label{subfig-1:dummy}]{%
      \includegraphics[width=0.13\textwidth]{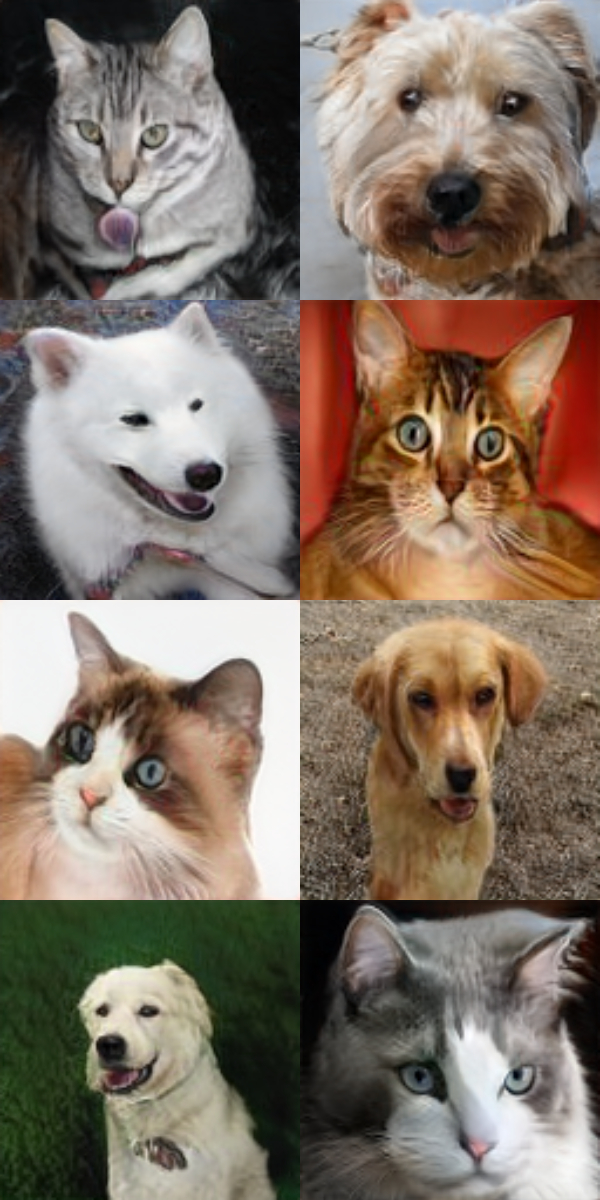}
    }
    \hfill\subfloat[\label{subfig-1:dummy}]{%
      \includegraphics[width=0.13\textwidth]{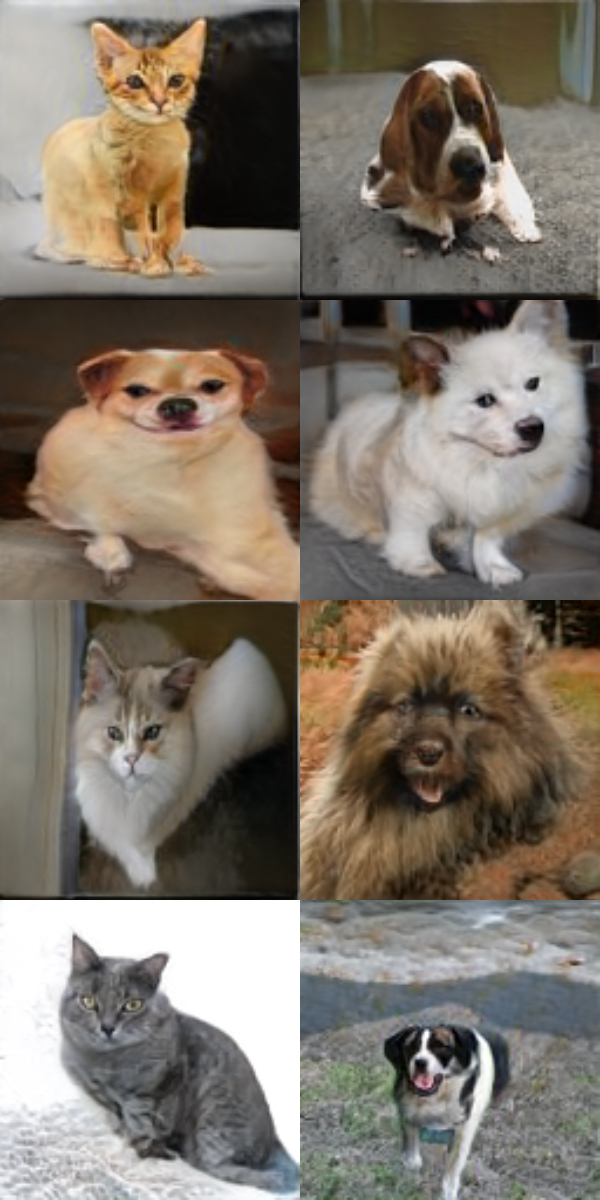}
    }
    \hfill\subfloat[\label{subfig-1:dummy}]{%
      \includegraphics[width=0.13\textwidth]{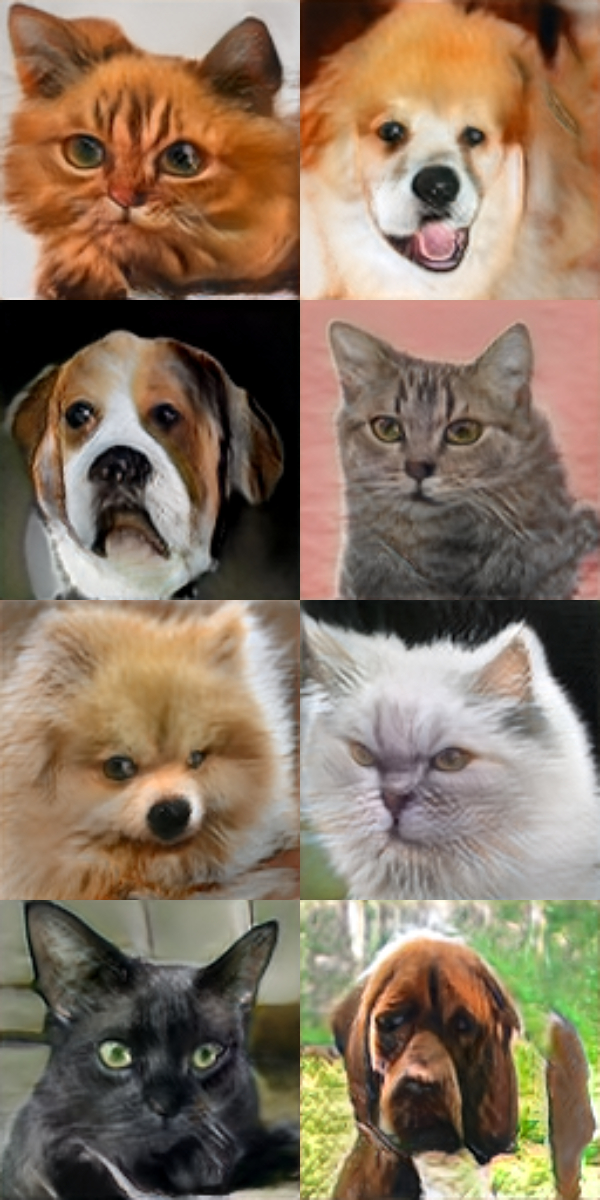}
    }
    \hfill\subfloat[\label{subfig-1:dummy}]{%
      \includegraphics[width=0.13\textwidth]{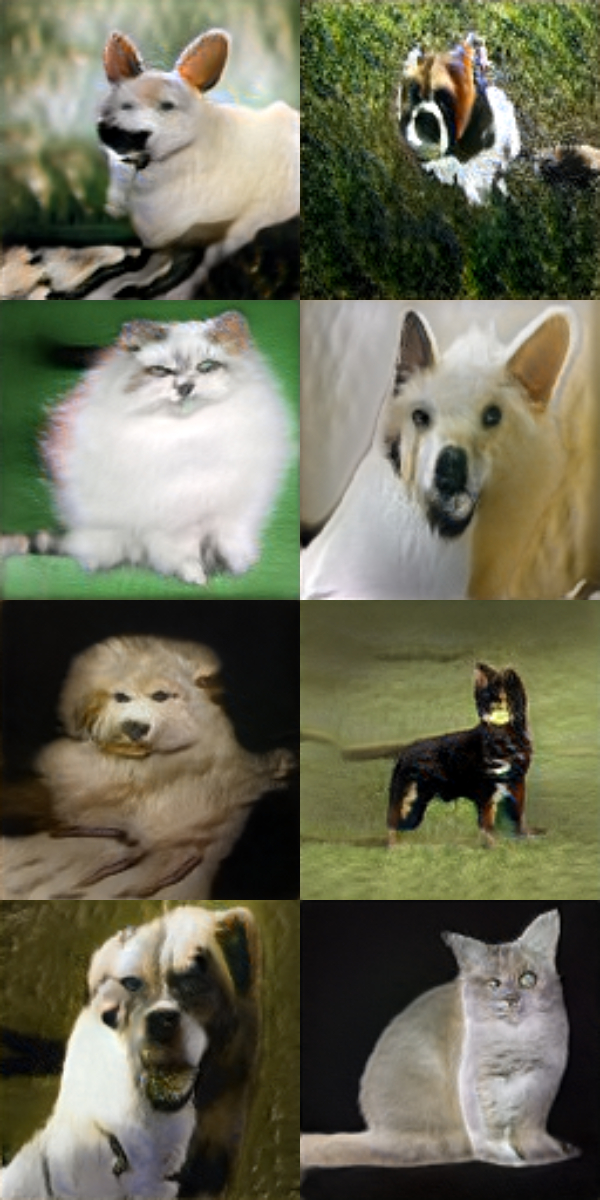}
    }
    \caption{Comparison between synthetic and authentic images. We show (a) the original data,(b) and (c) generated images on the whole dataset, cropped and uncropped, respectively. (d) cropped images on 50\%, (e) uncropped images generated on 50\% subset and finally (f) and (g), cropped and uncropped images result of training on only 10\% of the data. These qualitative visualizations prove the effectiveness and the interpretability of our method.}
      \label{fig:gan}
  \end{figure*}

Our procedure consists of using GANs to expand the length of the dataset by generating new instances of the training set (see examples in Figure \textcolor{red}{\ref{fig:gan}}). We further improved the generated images' quality by detecting the facial landmarks and cropping images fed to the GANs model. Moreover, we compared the result with original data. Finally, by combining synthetic images with real ones, we trained on the highly developed classifier model to show the effectiveness of our method, The whole pipeline is illustrated in Figure \textcolor{red}{\ref{fig:pipe}}.

\subsection{Fine-Grained Classification Model}

Vision transformers recently achieved exceptional results on various datasets. ViT \cite{dosovitskiy2021image} and BiT \cite{kolesnikov2020big}, two popular transformer models, delivered outstanding performance on the Oxford-IIIT Pet dataset. In this work, we train our augmented dataset plus the original one on a variant of ViT named the R26 + ViT-S/32 model. This hybrid model is a ViT-S/32 on top of a ResNet-26 \cite{AAAI1714806} backbone; this combination allows us to achieve similar outcomes to larger models with less than half the computational finetuning cost. R26 + ViT-S/32 is pre-trained on ImageNet-21k and has fewer parameters comparing to larger variants of ViT. We chose this model due to limitations in computational power, even though models with more parameters yield better results. We compared the results and our method's capability by training each dataset variation, namely, original, augmented, cropped-augmented on the classifier model.

\begin{figure}[ht]
    \centering
	\includegraphics[clip,width=0.9\linewidth]{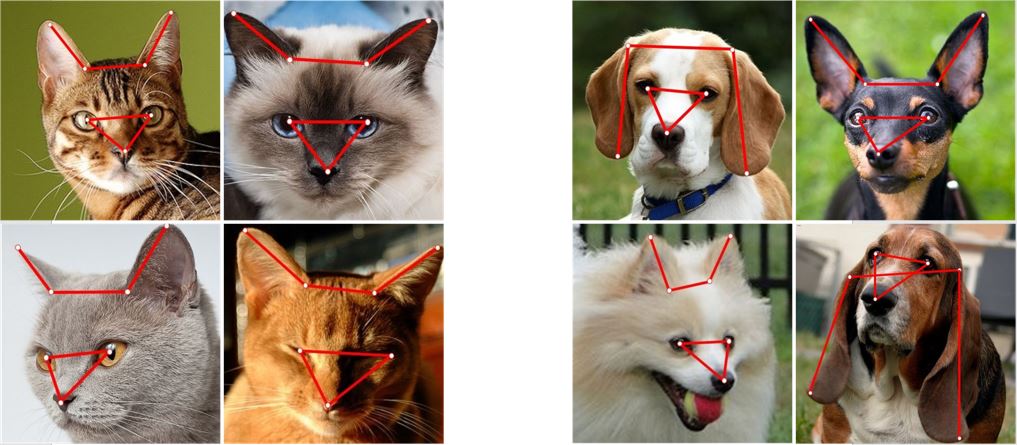}
	\caption{Examples of different cat and dog breeds and their associated landmarks}
		\label{fig:land}       
\end{figure}

\subsection{Data Augmentation via Image Generation}

By employing GANs framework to generate new images, it is possible to use them for the task of augmentation to reduce overfitting and increase the accuracy of the classifier. To keep the data balance, we used a conditional form of StyleGAN2-ADA, which receives an extra parameter as input $class = \{1, 2, ..., 37\}$ which denotes one of the 37 breeds available in the pets dataset. The conditional model works more efficiently than training each class separately. It consolidates data from different classes while keeping each of them realistic and does not confuse the looks of each breed together.

StyleGAN2-ADA takes advantage of augmentation to create notable results even with limited data. To experiment with the size of the dataset, we trained both StyleGAN and ViT models using $\{10\%, 50\%, 100\% \}$ subsets of pets dataset to show the effectiveness of dataset size in the output.

\subsection{Facial Landmark Detection}
The position and size of the object in the image ( in this case, cats and dogs) play a crucial role in robust image generation using GANs. Most publicly available datasets do not meet this requirement; thus, GANs are limited to a petite number of datasets, i.e., CelebA \cite{liu2015faceattributes}, AFHQ \cite{choi2020starganv2}, etc. In this paper, we demonstrate how preparing the Oxford-IIIT Pet dataset by cropping the images can improve the performance of GANs and therefore generate more realistic images. Due to inconsistent annotations in the pets dataset, instead of using provided bounding boxes, we trained a model to predict the animal landmarks, i.e., ears, nose, and eyes. Furthermore, we cropped the images by taking advantage of predicted landmarks. This method can easily generalize on different datasets, despite the bounding box restrictions.

We finetuned a MoblieNetV2 model on annotated datasets from Sandler et al. \cite{Sandler_2018_CVPR}. The mentioned model was pre-trained on ImageNet, which outputs 14 numbers denoting 7 landmarks on the animal face (see Figure  \textcolor{red}{\ref{fig:land}}). All the data is augmented with traditional methods, i.e., mirroring, rotation, and zooming. Furthermore, we used landmark normalization based on the center of the anchor, introduced in \cite{deng2019retinaface}, which slightly boosted the performance.

\section{Experiments}
\label{sec:exper}
\subsection{Datasets}

\textbf{Oxford IIIT-Pet Dataset} To demonstrate the capability of our proposed method, we chose the public Oxford-IIIT Pet \cite{6248092} dataset, which includes pictures of animals in different shapes and poses. This diversity suggests a more natural distribution and intensifies our work's performance compared to the benchmark results. This dataset contains a total of 7349 images in 37 different breeds of cats (12 categories) and dogs (25 categories). Every species consists of roughly 200 photos in JPG format and resized to 128 $\times$ 128 for training StyleGAN2-ADA, and fine-tuning ViT.

We used the official training split according to the original paper for both training the GANs model as well as fine tuning the fine-grained image classifier. To demonstrate the effect of the proposed augmentation technique, we performed all experiments on subsets of the dataset, i.e., 10\%, 50\%, and 100\%, both on cropped and the original data.

\textbf{Landmark Detection Datasets} We benefit from of two separate datasets to finetune a MobileNet model for the purpose of facial landmark prediction of cats and dogs. By combining the dog face dataset collected by Mougeot et al. \cite{10.1007/978-3-030-29894-4_34}. with a variation of the cat face dataset from Zhang et al. \cite{weiweizhangandjiansunandxiaooutang2008cat}, we produced a consolidated dataset for the mentioned task. In addition, we unified and improved the landmarks format. The first dataset contains 1393 classes of dogs, 8363 images, and there are at least two images per dog. Each image is of size 224 $\times$ 224 $\times$ 3 and has JPG format. The second dataset is consists of 10,000 annotated cat images collected from \textit{flickr.com}. To keep the unification of both datasets, we resized all images from the cat dataset to 224 $\times$ 224. Lastly, we used the official train/validation/test split suggested in the paper.

\subsection{Implementation Details}

\textbf{MobileNetV2 Settings} We used original MobileNetV2 architecture with some modifications for facial landmark prediction. It accepts 224 $\times$ 224 RGB images as input. Global max-pooling is applied to pooling layers. We discarded the Fully Connected Layers (FCL) on top; instead, we used two FCL with $2^7$ nodes and Rectified Linear Unit (ReLu) activation function. The model's output is the size of 14, which denotes 7 coordinates of the animal landmark (2 for each ear, 1 for each eye, 1 for nose).

\textbf{StyleGAN2-ADA Settings} This model has trained six times on different variants of the pets dataset, with the same hyperparameters. All input images are 128 $\times$ 128 $\times$ 3, and the conditional form of the model is used with 37 classes for each species type. ADA target value is set to 0.6, which controls data augmentation intensity and, methods of augmentation are set to default. Learning rates for generator and discriminator are both set to 0.0025.

\textbf{R26  +  ViT-S/32 Settings}  To keep the results comparable with the available performance of the ViT model, we used the same hyperparameters as mentioned in the official paper; however, because of the lower resolution of generated images, we resized the final dataset to 128 $\times$ 128 for consistency. The model is pre-trained on ImageNet with 85.99\% accuracy, and dropout is excluded. Additionally, Adam optimizer performs slightly better than Stochastic Gradient Descent (SGD). We split the data and evaluated the model using 3680 images for training and 3669 images for testing.

\subsection{Training Details}
Our model is implemented using PyTorch on an Ubuntu
20.04 server. We use a Tesla P100 GPU to accelerate
our training process. We finetuned MobileNetV2 model for 15 epochs and 32 batch size. As recommended in StyleGAN2-ADA paper, 5000 kimg training results in realistic images and it is almost converged. We trained all variants from scratch up to 5120 kimg, each of them took approximately 60 hours. Finally, ViT model pre-trained for 300 epochs, then fined tuned for 16 more on each variant of pets dataset.

\begin{table}[t]
\centering
\begin{tabular}{p{0.30\linewidth}m{0.10\linewidth}m{0.10\linewidth}}
\toprule
\multirow{2}{*}{Models} & \multicolumn{2}{c}{RMSE ($\downarrow$)}\\ \cmidrule{2-3}
 &   Validation &   Test \\ 
\midrule 
MobileNetV2 & 3.11 & 3.56 \\ \addlinespace
MobileNetV2 +Normalization & 3.02 & 3.43 \\ 
\bottomrule
\end{tabular}
\caption{We evaluate the RMSE once for the MobileNetV2 model and once for MobileNetV2 that is added with Normalization. The RMSE is calculated for both validation and test set.}
\label{table:one}
\end{table}

\pgfplotstableread{graph.dat}{\loadedtable}
 
  \begin{figure*}
  \centering
  
\subfloat{
\begin{tikzpicture}
\begin{axis}[
    title = 100\% subset,
    xlabel = $\times 10^3 \quad kimg$,
    ylabel = FID,
    xmin = 0, xmax = 5.04,
    ymin = 10, ymax = 75,
    xtick distance = 1,
    ytick distance = 20,
    grid = both,
    minor tick num = 1,
    major grid style = {lightgray},
    minor grid style = {lightgray!25},
    width = 0.3\textwidth,
]
\addplot[blue] table [x = {x}, y = {y3}] {\loadedtable};
\addplot[red] table [x ={x}, y = {y4}] {\loadedtable};
\end{axis}
\end{tikzpicture}
}\hfill
\subfloat{
\begin{tikzpicture}
\begin{axis}[
    legend columns=-1,
    legend entries={cropped,uncropped},
    legend to name=named,
    title = 50\% subset,
    xlabel = $\times 10^3 \quad kimg$,
    ylabel = FID,
    xmin = 0, xmax = 5.04,
    ymin = 15, ymax = 100,
    xtick distance = 1,
    ytick distance = 20,
    grid = both,
    minor tick num = 1,
    major grid style = {lightgray},
    minor grid style = {lightgray!25},
    width = 0.3\textwidth,]
\addplot[blue] table [x = {x}, y = {y2}] {\loadedtable};
\addplot[red] table [x ={x}, y = {y5}] {\loadedtable};
\end{axis}
\end{tikzpicture}
} \hfill
\subfloat{
\begin{tikzpicture}
\begin{axis}[
    title = 10\% subset,
    xlabel = $\times 10^3 \quad kimg$,
    ylabel = $FID$,
    xmin = 0, xmax = 5.04,
    ymin = 40, ymax = 150,
    xtick distance = 1,
    ytick distance = 25,
    grid = both,
    minor tick num = 1,
    major grid style = {lightgray},
    minor grid style = {lightgray!25},
    width =0.3\textwidth,
]
\addplot[blue] table [x = {x}, y = {y1}] {\loadedtable};
\addplot[red] table [x ={x}, y = {y6}] {\loadedtable};
\end{axis}
\end{tikzpicture}
} \\
\textcolor{black}{\ref{named}}
\caption{We evaluate the accuracy of the used model and FID for three different dataset conditions (Original, augmented, and cropped-augmented ) in data regimes of 10, 50, and 100 percent.}
\label{fig:fid}
\end{figure*}
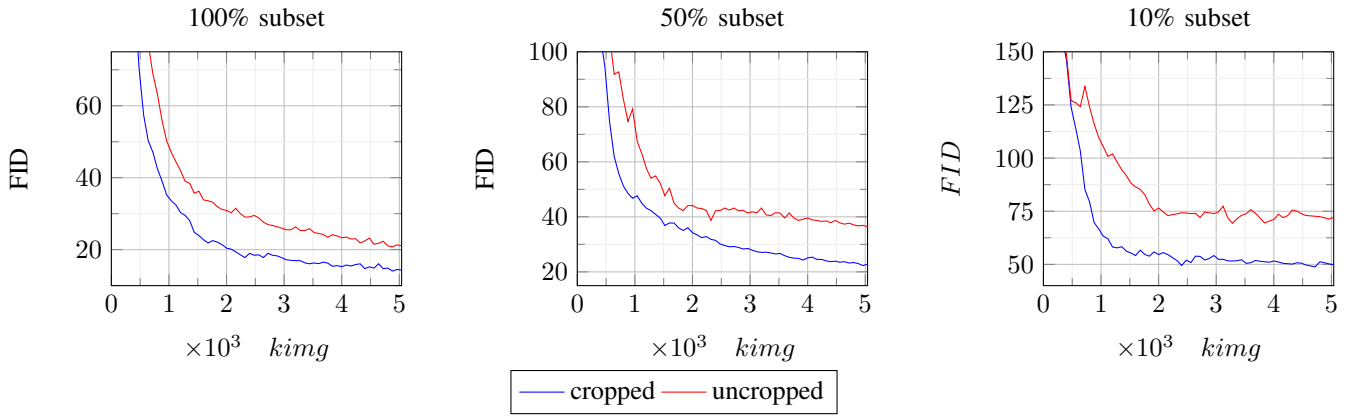
\begin{table*}[ht]
\centering
\setlength{\tabcolsep}{10pt}
\begin{tabular}{p{0.29\textwidth}*{4}{l{c}}}
\toprule
\multicolumn{1}{l}{\multirow{2}{*}{ Dataset variant}} & \multicolumn{2}{c}{10\% training data} & \multicolumn{2}{c}{50\% training data} & \multicolumn{2}{c}{100\% training data} \\
\cmidrule(r{3.8pt}){2-3} \cmidrule(l){4-5} \cmidrule(l){6-7}
& FID($\downarrow$) &   Test Accuracy ($\uparrow$)  &  FID &   Test Accuracy &  FID &   Test Accuracy  \\
\midrule
Original                & --    & 64.73 & --    & 88.41 & --     & 94.13\\\addlinespace
Augmented               &  71.1    & 63.32   & 36.4 & 88.70 & 20.7 & 94.93\\\addlinespace
Cropped-Augmented (Ours)& \textbf{49.4} & \textbf{68.55}  & \textbf{22.3} & \textbf{91.73} & \textbf{14.1} & \textbf{96.28} \\

\bottomrule
\end{tabular}
\caption{We evaluate the accuracy of the used model and FID for three different dataset conditions (Original, augmented, and augmented-cropped ) in data regimes of 10, 50, and 100 percent.}
\label{table:two}
\end{table*}

\section{Results}
\label{sec:result}
\subsection{Evaluation Metrics}

We used three independent metrics to evaluate the performances of our approaches and tasks. For the fine-grained image classification task, we considered the Accuracy metric since there are pretty equal samples belonging to each class. Accuracy has the following definition where $Ra$ is the total number of input samples and $R$ means the number of testing images.
 $$
  Accuracy  = \frac{R_a}{R}
 $$

In order to assess the quality of the images that
are generated, Frechet Inception Distance (FID) is
used. The lower amount of FID displays better quality.
In FID, first, a pre-trained InceptionV3 model is loaded and removes the features from an intermediate layer. Then the output is taken as the activations from the last pooling layer. A multivariate Gaussian distribution is used to model the data arrangement using covariance $\Sigma$ and mean $\mu$.
The FID interpolated the authentic images $x$ and produced
images $g$ is calculated as follows, and $Tr$ sums up all the diagonal elements:

 $$
  FID(x, g) = {||\mu_x - \mu_g||_2}^2 + Tr(\Sigma_x + \Sigma_g - 2\sqrt{\Sigma_x \times\Sigma_g })
  $$

For measuring facial landmarking accuracy, we used the Root Mean Square Error (RMSE) \cite{Johnston2018}, which is a metric that shows the average distance between the predicted values from the model and the actual values in the dataset. We used the following formula to measure the average length between every $N$ anticipated landmarks $ (x_i^p, y_i^p) $ and the corresponding 'ground truth' $ (x_i^t, y_i^t) $ on a per landmark basis. The predicted landmarks that aren't good enough will be placed far-off from their related provided annotations positions and cause to raise the value of RMSE.
$$
 RMSE = \frac{1}{N}\sum_{i=1}^{N} \sqrt{(x_i^p - x_i^t)^2 + (y_i^p - y_i^t)^2}
$$

\subsection{Performance Analysis}

To evaluate the effectiveness of our landmark predictor, we evaluated RMSE of the trained MobileNetV2 model with and without landmark normalization. Table  \textcolor{red}{\ref{table:one}} shows that normalization decreases RMSE on both validation and test set, suggesting increased accuracy and quality in predicted landmarks. Since we detect only one object per image, in some cases where multiple instances of dogs or cats were visible, predicting the landmark for the wrong instance caused the introduction of outliers. We removed these outliers to keep the RMSE more legitimate.

The reported results in Table  \textcolor{red}{\ref{table:two}} show that both centering the objects and manipulating the size of the dataset have a discernible effect on the quality of generated outputs, measured by their FID. Cropping the images decreased FID substantially on all three variants of the dataset. On the 100\% subset of the cropped dataset, FID reached as low as 15.9, which is 32\% less than the uncropped version. Limiting the amount of data fed into the GANs model had a denying impact on the condition of generated images that are shown in Figure \textcolor{red}{\ref{fig:gan}} for variants of the dataset. Regarding Figure \textcolor{red}{\ref{fig:fid}}, it is noticeable that FID is still slightly decreasing and have not converged yet, but due to computational constraints, we stopped at approximately 5000 kimg.

As shown in Table \textcolor{red}{\ref{table:two}}, the size of the database has a direct impact on the accuracy of experimented ViT model. It is clear that data augmentation has a negligible influence on the results when the dataset is not prepared and even reduced accuracy on the 10\% subset due to the low quality of generated images and lack of detail. Our method generated higher quality instances and dealt with the problem of abstraction in images. Consequently, increased accuracy in all three subsets of the dataset and improved the available benchmark by roughly 2\%.
        
\section{conclusion}
\label{sec:con}
      In this work,  we gained improvement in fine-grained image classification by generating new samples with the use of GANs. New data samples were obtained from StyleGAN2-ADA. We finetuned a custom MobileNetV2 model to predict animal facial landmarks, then cropped the Oxford-IIIT Pet dataset images accordingly. The newly generated images from the cropped dataset caused to enhancement in quality and diversity. An increase in the number and diversification of sample images impacts increasing the accuracy of the state-of-the-art ViT model.

\bibliographystyle{ieeetr}
\bibliography{sample}

\end{document}